\documentclass[11pt,a4paper]{article}
\usepackage[hyperref]{acl2021}
\usepackage{times}
\usepackage{latexsym}
\usepackage{amsmath}
\usepackage{graphicx}
\usepackage{caption}
\usepackage{subcaption}
\usepackage{enumitem}
\usepackage{tabularx}

\usepackage{lipsum} 
\usepackage{multirow}
\usepackage{hyperref}

\newcommand*\rot{\rotatebox[origin=c]{90}}

\usepackage{microtype}

\aclfinalcopy 
\def\aclpaperid{3578} 

\title{Self-Supervised Multimodal Opinion Summarization}

\author{Jinbae Im, Moonki Kim, Hoyeop Lee, Hyunsouk Cho, Sehee Chung
 \\
  Knowledge AI Lab., NCSOFT Co., South Korea \\
  \texttt{\{jinbae,kmkkrk,hoyeoplee,dakgalbi,seheechung\}@ncsoft.com}}

\date{}

\begin{document}
\maketitle
\begin{abstract}

Recently, opinion summarization, which is the generation of a summary from multiple reviews, has been conducted in a self-supervised manner by considering a sampled review as a pseudo summary. However, non-text data such as image and metadata related to reviews have been considered less often. To use the abundant information contained in non-text data, we propose a self-supervised multimodal opinion summarization framework called MultimodalSum. Our framework obtains a representation of each modality using a separate encoder for each modality, and the text decoder generates a summary. To resolve the inherent heterogeneity of multimodal data, we propose a multimodal training pipeline. We first pretrain the text encoder--decoder based solely on text modality data. Subsequently, we pretrain the non-text modality encoders by considering the pretrained text decoder as a pivot for the homogeneous representation of multimodal data. Finally, to fuse multimodal representations, we train the entire framework in an end-to-end manner. We demonstrate the superiority of MultimodalSum by conducting experiments on Yelp and Amazon datasets. 
\end{abstract}

\section{Introduction}

\begin{figure}[t!]
     \centering
     \begin{subfigure}{0.48\textwidth}
         \centering
         \includegraphics[width=\textwidth]{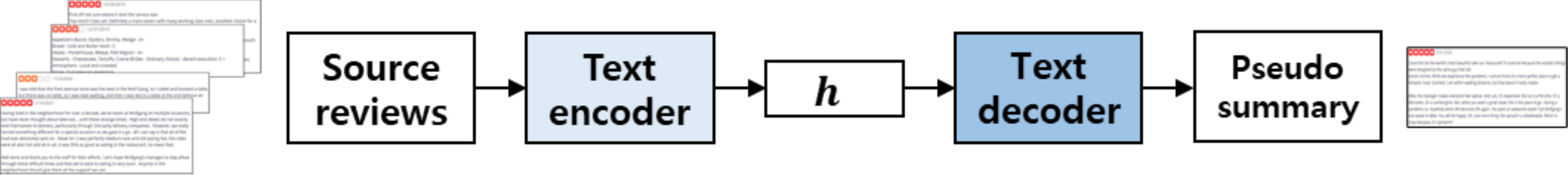}
         \caption{Unimodal framework}
         \label{fig:main1}
         \vspace{2mm}
     \end{subfigure}
     \begin{subfigure}{0.48\textwidth}
         \centering
         \includegraphics[width=\textwidth]{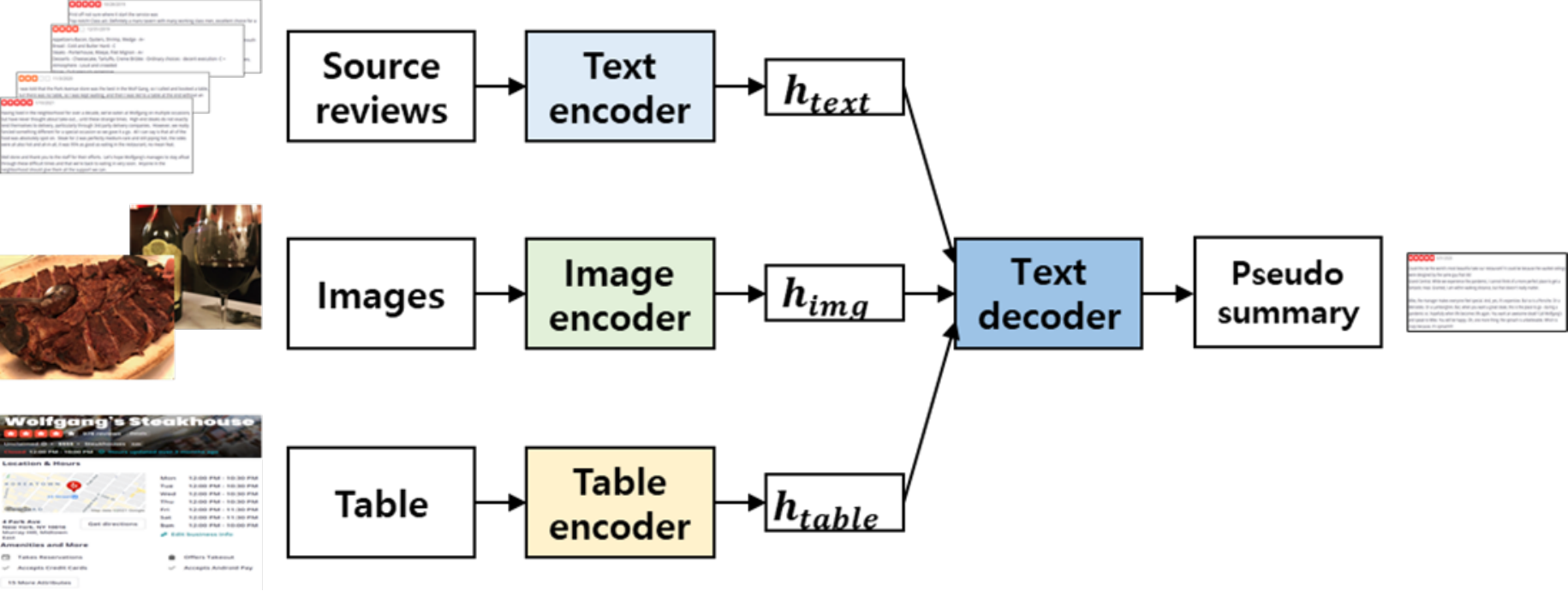}
         \caption{Multimodal framework}
         \label{fig:main2}
     \end{subfigure}
     \caption{Self-supervised opinion summarization frameworks}
     \label{fig:main}
\end{figure}

Opinion summarization is the task of automatically generating summaries from multiple documents containing users' thoughts on businesses or products. This summarization of users' opinions can provide information that helps other users with their decision-making on consumption. Unlike conventional single-document or multiple-document summarization, where we can obtain the prevalent annotated summaries~\citep{Nallapati2016Abstractive, See2017Get, Paulus2018A, Liu2018Generating, Liu2019Hierarchical, Perez-Beltrachini2019Generating}, opinion summarization is challenging; it is difficult to find summarized opinions of users. Accordingly, studies used an unsupervised approach for opinion summarization~\citep{Ku2006Opinion, Paul2010Summarizing, Carenini2013Multi, Ganesan2010Opinosis, Gerani2014Abstractive}. Recent studies~\citep{Brazinskas2020Unsupervised, Amplayo2020Unsupervised, Elsahar2020Self} used a self-supervised learning framework that creates a synthetic pair of source reviews and a pseudo summary by sampling a review text from a training corpus and considering it as a pseudo summary, as in Figure~\ref{fig:main1}.

Users' opinions are based on their perception of a specific entity and perceptions originate from various characteristics of the entity; therefore, opinion summarization can use such characteristics.
For instance, Yelp provides users food or menu images and various metadata about restaurants, as in Figure~\ref{fig:main2}.
This non-text information influences the review text generation process of users~\citep{Truong2019Multimodal}.
Therefore, using this additional information can help in opinion summarization, especially under unsupervised settings~\citep{Su2019Unsupervised, Huang2020Unsupervised}.
Furthermore, the training process of generating a review text (a pseudo summary) based on the images and metadata for self-supervised learning is consistent with the actual process of writing a review text by a user.

This study proposes a self-supervised multimodal opinion summarization framework called MultimodalSum by extending the existing self-supervised opinion summarization framework, as shown in Figure~\ref{fig:main}.
Our framework receives source reviews, images, and a table on the specific business or product as input and generates a pseudo summary as output.
Note that images and the table are not aligned with an individual review in the framework, but they correspond to the specific entity.
We adopt the encoder--decoder framework and build multiple encoders representing each input modality.
However, a fundamental challenge lies in the heterogeneous data of various modalities~\citep{Baltruvsaitis2018Multimodal}.

To address this challenge, we propose a multimodal training pipeline.
The pipeline regards the text modality as a pivot modality. 
Therefore, we pretrain the text modality encoder and decoder for a specific business or product via the self-supervised opinion summarization framework.
Subsequently, we pretrain modality encoders for images and a table to generate review texts belonging to the same business or product using the pretrained text decoder.
When pretraining the non-text modality encoders, the pretrained text decoder is frozen so that the image and table modality encoders obtain homogeneous representations with the pretrained text encoder.
Finally, after pretraining input modalities, we train the entire model in an end-to-end manner to combine multimodal information.

Our contributions can be summarized as follows:
\begin{itemize}[noitemsep,topsep=0pt]
\item this study is the first work on self-supervised multimodal opinion summarization;
\item we propose a multimodal training pipeline to resolve the heterogeneity between input modalities;
\item we verify the effectiveness of our model framework and model training pipeline through various experiments on Yelp and Amazon datasets.
\end{itemize}

\section{Related Work}
Generally, opinion summarization has been conducted in an unsupervised manner, which can be divided into extractive and abstractive approaches. 
The extractive approach selects the most meaningful texts from input opinion documents, and the abstractive approach generates summarized texts that are not shown in the input documents.
Most previous works on unsupervised opinion summarization have focused on extractive approaches.
Clustering-based approaches~\citep{Carenini2006Multi, Ku2006Opinion, Paul2010Summarizing, Angelidis2018Summarizing} were used to cluster opinions regarding the same aspect and extract the text representing each cluster. 
Graph-based approaches~\citep{Erkan2004Lexrank, Mihalcea2004Textrank, Zheng2019Sentence} were used to construct a graph---where nodes were sentences, and edges were similarities between sentences---and extract the sentences based on their centrality. 

Although some abstractive approaches were not based on neural networks~\citep{Ganesan2010Opinosis, Gerani2014Abstractive, Di2014Hybrid}, neural network-based approaches have been gaining attention recently. \citet{Chu2019Meansum} generated an abstractive summary from a denoising autoencoder-based model. 
More recent abstractive approaches have focused on self-supervised learning.
\citet{Brazinskas2020Unsupervised} randomly selected $N$ review texts for each entity and constructed $N$ synthetic pairs by sequentially regarding one review text as a pseudo summary and the others as source reviews.
\citet{Amplayo2020Unsupervised} sampled a review text as a pseudo summary and generated various noisy versions of it as source reviews.
\citet{Elsahar2020Self} selected review texts similar to the sampled pseudo summary as source reviews, based on TF-IDF cosine similarity.
We construct synthetic pairs based on \citet{Brazinskas2020Unsupervised} and extend the self-supervised opinion summarization to a multimodal version.

Multimodal text summarization has been mainly studied in a supervised manner.
Text summaries were created by using other modality data as additional input~\citep{Li2018Multi,Li2020Aspect}, and some studies provided not only a text summary but also other modality information as output~\citep{Zhu2018MSMO,Chen2018Abstractive,Zhu2020Multimodal,Li2020VMSMO,Fu2020Multi}. 
Furthermore, most studies summarized a single sentence or document. Although \citet{Li2020Aspect} summarized multiple documents, they used non-subjective documents. Our study is the first unsupervised multimodal text summarization work that summarizes multiple subjective documents.

\section{Problem Formulation}

The goal of the self-supervised multimodal opinion summarization is to generate a pseudo summary from multimodal data. Following existing self-supervised opinion summarization studies, we consider a review text selected from an entire review corpus as a pseudo summary. We extend the formulation of \citet{Brazinskas2020Unsupervised} to a multimodal version. Let $R$ = $\{r_1, r_2, ..., r_N\}$ denote the set of reviews about an entity (e.g., a business or product). Each review, $r_j$, consists of review text, $d_j$, and review rating, $s_j$, that represents the overall sentiment of the review text. We denote images uploaded by a user or provided by a company for the entity as $I$ = $\{i_1, i_2, ..., i_M\}$ and a table containing abundant metadata about the entity as $T$. Here, $T$ consists of several fields, and each field contains its own name and value. We set $j$-th review text $d_j$ as the pseudo summary and let it be generated from $R_{-j}$, $I$, and $T$, where $R_{-j}=\{r_1, ..., r_{j-1}, r_{j+1}, ..., r_N\}$ denotes source reviews. To help the model summarize what stands out overall in the review corpus, we calculate the loss for all $N$ cases of selecting $d_j$ from $R$, and train the model using the average loss. During testing, we generate a summary from $R$, $I$, and $T$.

\section{Model Framework}

The proposed model framework, MultimodalSum, is designed with an encoder--decoder structure, as in Figure~\ref{fig:main2}.
To address the heterogeneity of three input modalities, we configure each modality encoder to effectively process data in each modality. We set a text decoder to generate summary text by synthesizing encoded representations from the three modality encoders. Details are described in the following subsections.

\begin{figure*}
    \centering
    \begin{minipage}{.45\linewidth}
        \begin{subfigure}[t]{\linewidth}
            \includegraphics[width=\textwidth]{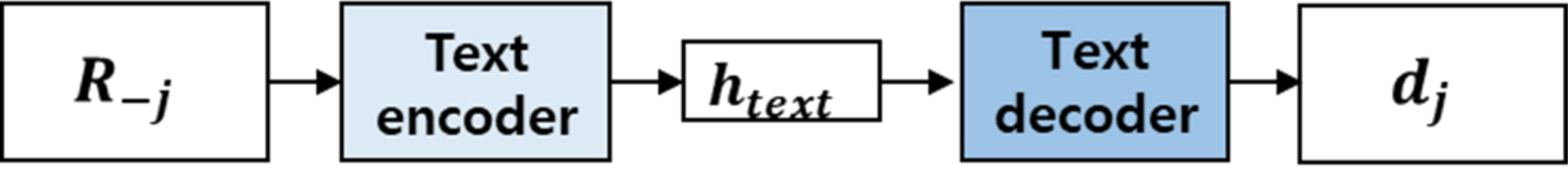}
            \caption{Text modality pretraining}
            \label{fig:fig_text}
        \end{subfigure} \\
        \begin{subfigure}[b]{\linewidth}
            \vspace{1.5mm}
            \includegraphics[width=\textwidth]{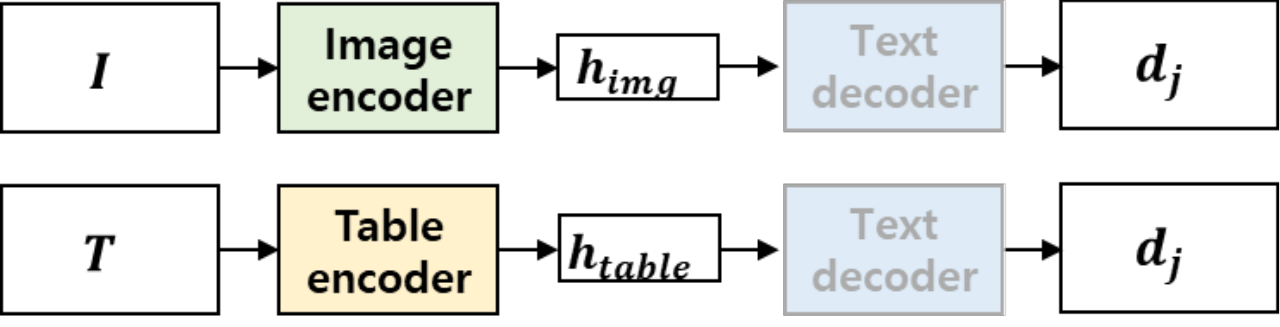}
            \caption{Other modalities pretraining}
            \label{fig:fig_other}
        \end{subfigure}
    \end{minipage}
    \qquad
    \begin{minipage}{.45\textwidth}
        \begin{subfigure}[t]{\linewidth}
            \includegraphics[width=\textwidth]{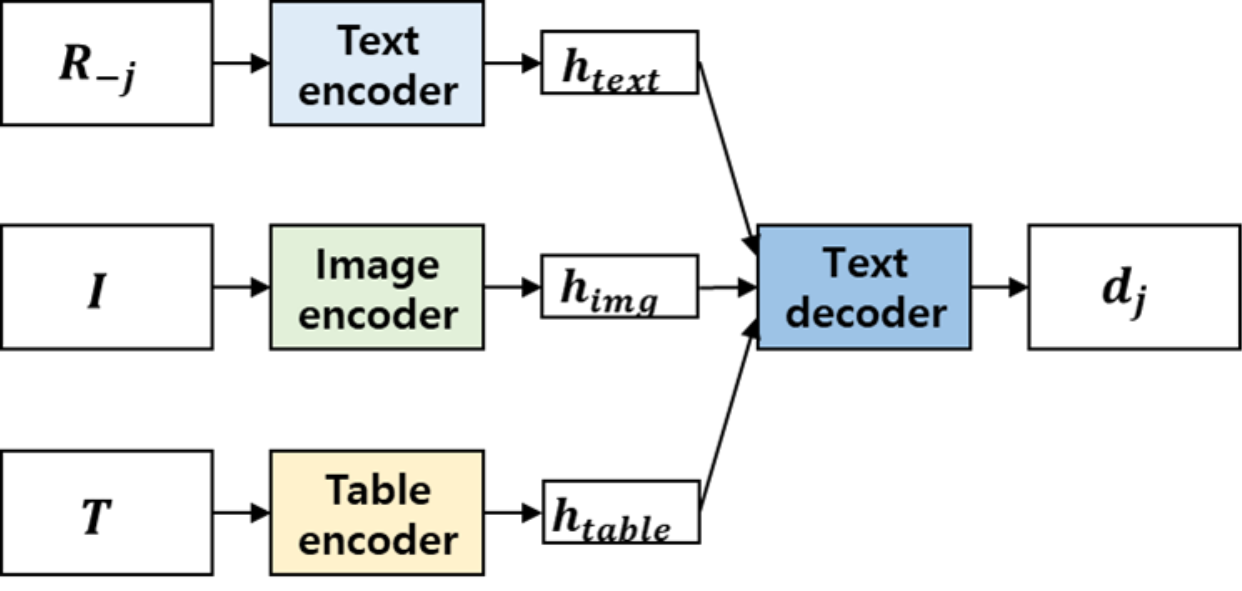}
            \caption{Training for multiple modalities}
            \label{fig:fig_multi}
        \end{subfigure}
    \end{minipage}

\caption{Self-supervised multimodal opinion summarization training pipeline. Blurred boxes in ``Other modalities pretraining'' indicate that the text decoders are untrained.}
\label{fig:Pipeline}
\end{figure*}

\subsection{Text Encoder and Decoder}

Our text encoder and decoder are based on BART~\citep{Lewis2020BART}. 
BART is a Transformer~\citep{Vaswani2017Attention} encoder--decoder pretrained model that is particularly effective when fine-tuned for text generation and has high summarization performance.
Furthermore, because the pseudo summary of self-supervised multimodal opinion summarization is an individual review text ($d_j$), we determine that pretraining BART based on a denoising autoencoder is suitable for our framework. Therefore, we further pretrain BART using the entire training review corpus~\citep{Gururangan2020Don't}.
Our text encoder obtains $e_D$-dimensional encoded text representations $h_\mathrm{text}$ from $D_{-j}$ and the text decoder generates $d_j$ from $h_\mathrm{text}$ as follows:
\begin{equation*}
    \begin{aligned}
        h_\mathrm{text} &= \operatorname{BART_{enc}}(D_{-j}), \\
        d_j &= \operatorname{BART_{dec}}(h_\mathrm{text}),
    \end{aligned}
\end{equation*} 
where $D_{-j}=\{d_1, ..., d_{j-1}, d_{j+1}, ..., d_N\}$ denotes the set of review texts from $R_{-j}$. Each review text consists of $l_D$ tokens and $h_\mathrm{text} \in R^{{(N-1)}\times{l_D}\times{e_D}}$.

\subsection{Image Encoder}

We use a convolutional neural network specialized in analyzing visual imagery. In particular, we use ImageNet pretrained ResNet101~\citep{He2016Deep}, which is widely used as a backbone network. We add an additional linear layer in place of the image classification layer to match feature distribution and dimensionality with text modality representations.
Our image encoder obtains encoded image representations $h_\mathrm{img}$ from $I$ as follows:
\begin{equation*}
    \begin{aligned}
        h_\mathrm{img} &=\operatorname{ResNet101}(I)\,W_\mathrm{img},
    \end{aligned}
\end{equation*}
where $W_\mathrm{img} \in R^{e_I \times e_D}$ denotes the additional linear weights. $h_\mathrm{img}$ obtains $R^{M \times l_I \times e_D}$, where $l_I$ represents the size of the flattened image feature map obtained from ResNet101.

\subsection{Table Encoder}

To effectively encode metadata, we design our table encoder based on the framework of data-to-text research~\citep{Puduppully2019Data}. The input to our table encoder $T$ is a series of field-name and field-value pairs. Each field gets $e_T$-dimensional representations through a multilayer perceptron after concatenating the representations of field-name and field-value. The encoded table representations $h_\mathrm{table}$ is obtained by stacking each field representation into $F$ and adding a linear layer as follows: 
\begin{equation*}
    \begin{aligned}
        f_k &= \operatorname{ReLU}([n_k; v_k]\,W_f + b_f), \\
        h_\mathrm{table} &= F\,W_\mathrm{table},
    \end{aligned}
\end{equation*} 
where $n$ and $v$ denote $e_T$-dimensional representations of field name and value, respectively, and $W_f \in R^{2e_T \times e_T}$, $b_f \in R^{e_T}$ are parameters. By stacking $l_T$ field representations, we obtain $F \in R^{1 \times l_T \times e_T}$. The additional linear weights $W_{table} \in R^{e_T \times e_D}$ play the same role as in the image encoder, and $h_{table} \in R^{1 \times l_T \times e_D}$.

\section{Model Training Pipeline}

To effectively train the model framework, we set a model training pipeline, which consists of three steps, as in Figure~\ref{fig:Pipeline}. The first step is text modality pretraining, in which a model learns unsupervised summarization capabilities using only text modality data. Next, during the pretraining for other modalities, an encoder for each modality is trained using the text modality decoder learned in the previous step as a pivot. The main purpose of this step is that other modalities have representations whose distribution is similar to that of the text modality. In the last step, the entire model framework is trained using all the modality data. Details of each step can be found in the next subsections.

\subsection{Text Modality Pretraining}

In this step, we pretrain the text encoder and decoder for self-supervised opinion summarization.
As this was an important step for unsupervised multimodal neural machine translation~\citep{Su2019Unsupervised}, we apply it to our framework. 
For the set of reviews about an entity $R$, we train the model to generate a pseudo summary $d_j$ from source reviews $R_{-j}$ for all $N$ cases as follows: $\text{loss} = \sum_{j=1}^{N}{\text{log}\,p(d_j|R_{-j})}$.
The text encoder obtains $h_\mathrm{text} \in R^{{(N-1)} \times l_D \times e_D}$ from $D_{-j}$, and the text decoder aggregates the encoded representations of $N-1$ review texts to generate $d_j$. 
We model the aggregation of multiple encoded representations in the multi-head self-attention layer of the text decoder. To generate a pseudo summary that covers the overall contents of source reviews, we simply average the $N-1$ single-head attention results for each encoded representation ($R^{l_D \times e_D}$) at each head~\citep{Elsahar2020Self}.

\begin{figure}[t!]
     \centering
     \includegraphics[width=0.48\textwidth]{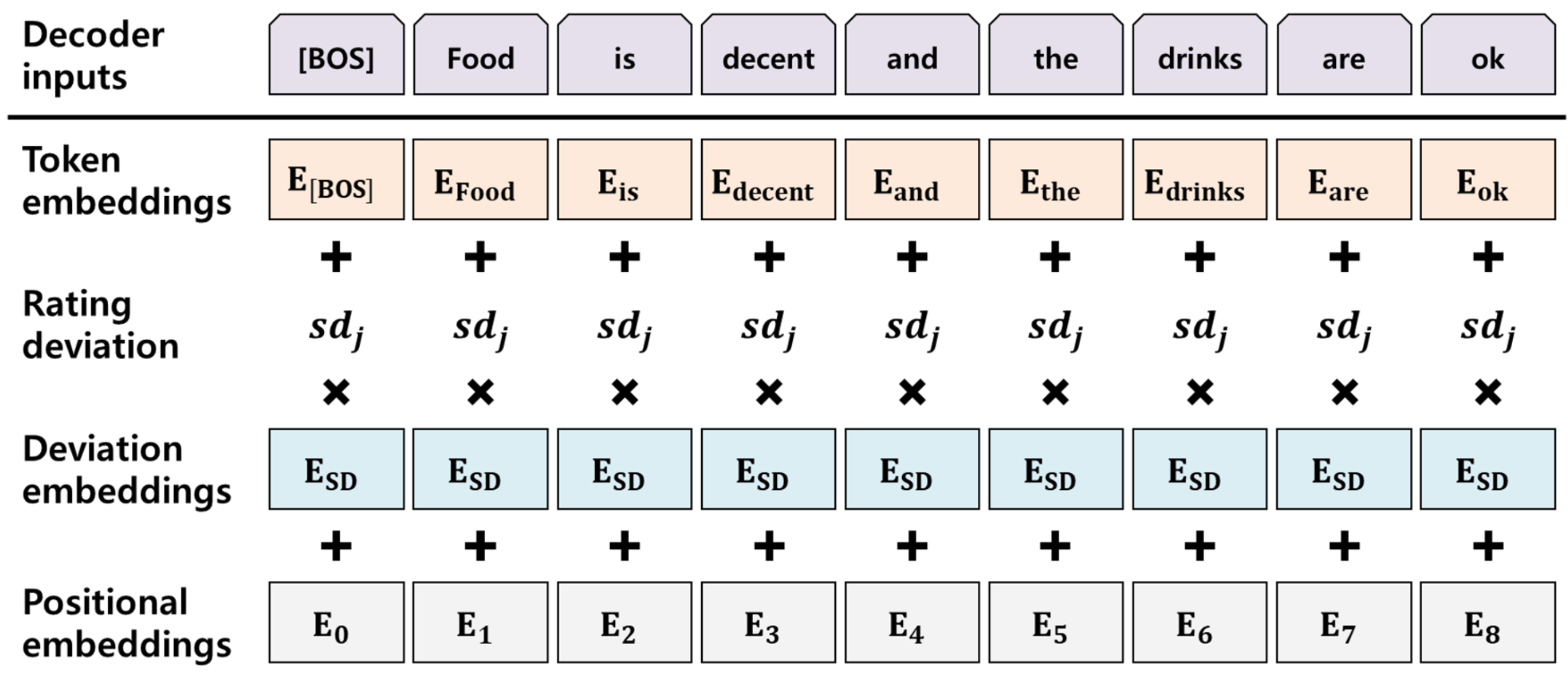}
     \caption{Text decoder input representations. The input embeddings are the sum of the token embeddings, rating deviation times deviation embeddings, and the positional embeddings.}
     \label{fig:deviation}
\end{figure}

The limitation of the self-supervised opinion summarization is that training and inference tasks are different. The model learns a review generation task using a review text as a pseudo summary; however, the model needs to perform a summary generation task at inference. To close this gap, we use a rating deviation between the source reviews and the target as an additional input feature of the text decoder, inspired by \citet{Bravzinskas2020Few}. We define the average ratings of the source reviews minus the rating of the target as the rating deviation: ${sd}_j = \sum_{i\neq j}^{N}{s_i} / (N-1) - s_j$. We use ${sd}_j$ to help generate a pseudo summary $d_j$ during training and set it as $0$ to generate a summary with average semantic of input reviews during inference. 
To reflect the rating deviation, we modify the way in which a Transformer creates input embeddings, as in Figure~\ref{fig:deviation}.
We create deviation embeddings with the same dimensionality as token embeddings and add ${sd}_j$ $\times$ deviation embeddings to the token embeddings in the same way as positional embeddings.

Our methods to close the gap between training and inference tasks do not require additional modeling or training in comparison with previous works.
We achieve noising and denoising effects by simply using rating deviation embeddings without variational inference in \citet{Brazinskas2020Unsupervised}. 
Furthermore, the information that the rating deviation is $0$ plays the role of an input prompt for inference, without the need to train a separate classifier for selecting control tokens to be used as input prompts~\citep{Elsahar2020Self}.

\subsection{Other Modalities Pretraining}

As the main modality for summarization is the text modality, we pretrain the image and table encoders by pivoting the text modality. Although the data of the three modalities are heterogeneous, each encoder should be trained to obtain homogeneous representations. We achieve this by using the pretrained text decoder as a pivot.
We train the image encoder and the table encoder along with the text decoder to generate a review text of the entity to which images or metadata belong: $I$ or $T$ $\rightarrow$ $d_j \in R$.
The image and table encoders obtain $h_\mathrm{img}$ and $h_\mathrm{table}$ from $I$ and $T$, respectively, and the text decoder generates $d_j$ from $h_\mathrm{img}$ or $h_\mathrm{table}$. Note that we aggregate $M$ encoded representations of $h_\mathrm{img}$ as in the text modality pretraining, and the weights of the text decoder are made constant.
$I$ or $T$ corresponds to all $N$ reviews, and this means that $I$ or $T$ has multiple references. We convert a multiple-reference setting to a single-reference setting to match the model output with the text modality pretraining. We simply create $N$ single reference pairs from each entity and shuffle pairs from all entities to construct the training dataset~\citep{Zheng2018Multi}. As the text decoder was trained for generating a review text from text encoded representations, the image and table encoders are bound to produce similar representations with the text encoder to generate the same review text. In this way, we can maximize the ability to extract the information necessary for generating the review text. 

\subsection{Training for Multiple Modalities}

We train the entire multimodal framework from the pretrained encoders and decoder.
The encoder of each modality obtains an encoded representation for each modality, and the text decoder generates the pseudo summary $d_j$ from multimodal encoded representations $h_\mathrm{text}$, $h_\mathrm{img}$, and $h_\mathrm{table}$.
To fuse multimodal representations, we aim to meet three requirements. First, the text modality, which is the main modality, is primarily used. Second, the model works even if images or metadata are not available. Third, the model makes the most of the legacy from pretraining.
To fulfill the requirements, multi-modality fusion is applied to the multi-head self-attention layer of the text decoder. 
The text decoder obtains the attention result for each modality at each layer.
We fuse the attention results for multiple modalities as follows:
\begin{equation*}
    \begin{aligned}
        {ma}_\mathrm{fused} = {ma}_\mathrm{text} + \alpha \odot {ma}_\mathrm{img} + \beta \odot {ma}_\mathrm{table},
    \end{aligned}
\end{equation*} 
where ${ma}_\mathrm{text}$, ${ma}_\mathrm{img}$, and ${ma}_\mathrm{table}$ denote each modality attention result from $h_\mathrm{text}$, $h_\mathrm{img}$, and $h_\mathrm{table}$, respectively.
$\odot$ symbolizes elementwise multiplication and $e_D$-dimensional multimodal gates $\alpha$ and $\beta$ are calculated as follows: $\alpha$ $=$ $\phi([ma_\mathrm{text}; ma_\mathrm{img}]\,W_{\alpha})$ and $\beta$ $=$ $\phi([ma_\mathrm{text}; ma_\mathrm{table}]\,W_{\beta})$. Note that $\alpha$ or $\beta$ obtains the zero vector when images or metadata do not exist.
It is common to use sigmoid as an activation function $\phi$. However, it can lead to confusion in the text decoder pretrained using only the text source. Because the values of $W$ are initialized at approximately $0$, the values of $\alpha$ and $\beta$ are initialized at approximately $0.5$ when sigmoid is used. To initialize the gate values at approximately $0$, we use ReLU(tanh($x$)) as $\phi(x)$.
This enables the continuous use of text information, and images or metadata are used selectively.

\section{Experimental Setup}

\begin{table}[t!]
\small
\begin{center}
\begin{tabular}{llccc}
\hline
Yelp                 & & Train\phantom{*}  & Dev   & Test        \\ \hline
\#businesses         & & 50,113\phantom{*} & 100   & 100         \\ 
\#reviews/business   & & 8\phantom{*}      & 8     & 8           \\
\#summaries/business & & 1*                & 1     & 1           \\
\#max images         & & 10\phantom{*}     & 10    & 10          \\
\#max fields         & & 47\phantom{*}     & 47    & 47          \\ \hline \hline
Amazon               & & Train\phantom{*}  & Dev   & Test        \\ \hline
\#products           & & 60,935\phantom{*} & 28    & 32          \\
\#reviews/product    & & 8\phantom{*}      & 8     & 8           \\
\#summaries/product  & & 1*                & 3     & 3           \\
\#max images         & & 1\phantom{*}      & 1     & 1           \\
\#max fields         & & 5+128\phantom{*}  & 5+128 & 5+128       \\ \hline
\end{tabular}
\end{center}
\vspace{-4mm}
\caption{Data statistics; $1$* in Train column indicates that it is a pseudo summary.}
\label{tab:data_statistics}
\end{table}

\subsection{Datasets}

To evaluate the effectiveness of the model framework and training pipeline on datasets with different domains and characteristics, we performed experiments on two review datasets: Yelp Dataset Challenge\footnote{https://www.yelp.com/dataset} and Amazon product reviews~\citep{He2016Ups}.
The Yelp dataset provides reviews based on personal experiences for a specific business. It also provides numerous images (e.g., food and drinks) uploaded by the users. Note that the maximum number of images, $M$, was set to $10$ based on the $90^{th}$ percentile.
In addition, the dataset contains abundant metadata of businesses according to the characteristics of each business.
On the contrary, the Amazon dataset provides reviews with more objective and specific details about a particular product. It contains a single image provided by the supplier, and provides relatively limited metadata for the product.  
For evaluation, we used the data used in previous research~\citep{Chu2019Meansum, Brazinskas2020Unsupervised}. The data were generated by Amazon Mechanical Turk workers who summarized $8$ input review texts. Therefore, we set $N$ to $9$ so that a pseudo summary is generated from $8$ source reviews during training. For the Amazon dataset, $3$ summaries are given per product.
Simple data statistics are shown in Table~\ref{tab:data_statistics}, and other details can be found in Appendix~\ref{appen:dataset}.

\subsection{Experimental Details}

All the models\footnote{Our code is available at \url{https://bit.ly/3bR4yod}} were implemented with PyTorch~\citep{Paszke2019Pytorch}, and we used the Transformers library from Hugging Face~\citep{Wolf2020Transformers} as the backbone skeleton. Our text encoder and decoder were initialized using BART-Large and further pretrained using the training review corpus with the same objective as BART. 
$e_D$, $e_I$, and $e_T$ were all set to 1,024.
We trained the entire models using the Adam optimizer~\citep{Kingma2014Adam} with a linear learning rate decay on NVIDIA V100s.
We decayed the model weights with $0.1$. For each training pipeline, we set different batch sizes, epochs, learning rates, and warmup steps according to the amount of learning required at each step.
We used label smoothing with $0.1$ and set the maximum norm of gradients as $1$ for other modalities pretraining and multiple-modalities training.
During testing, we used beam search with early stopping and discarded hypotheses that contain twice the same trigram. Different beam size, length penalty, and max length were set for Yelp and Amazon. The best hyperparameter values and other details are described in Appendix~\ref{appen:experimental}.

\subsection{Comparison Models}
We compared our model to extractive and abstractive opinion summarization models. For extractive models, we used some simple baseline models~\citep{Brazinskas2020Unsupervised}. \textit{Clustroid} selects one review that gets the highest ROUGE-L score with the other reviews of an entity. \textit{Lead} constructs a summary by extracting and concatenating the lead sentences from all review texts of an entity. \textit{Random} simply selects one random review from an entity. \textit{LexRank}~\citep{Erkan2004Lexrank} is an extractive model that selects the most salient sentences based on graph centrality. 

For abstractive models, we used non-neural and neural models. \textit{Opinosis}~\citep{Ganesan2010Opinosis} is a non-neural model that uses a graph-based summarizer based on token-level redundancy. \textit{MeanSum}~\citep{Chu2019Meansum} is a neural model that is based on a denoising-autoencoder and generates a summary from mean representations of source reviews. We also used three self-supervised abstractive models. \textit{DenoiseSum}~\citep{Amplayo2020Unsupervised} generates a summary by denoising source reviews. \textit{Copycat}~\citep{Brazinskas2020Unsupervised} uses a hierarchical variational autoencoder model and generates a summary from mean latent codes of the source reviews. \textit{Self \& Control}~\citep{Elsahar2020Self} generates a summary from Transformer models and uses some control tokens as additional inputs to the text decoder.

\section{Results}

We evaluated our model framework and model training pipeline.
In particular, we evaluated the summarization quality compared to other baseline models in terms of automatic and human evaluation, and conducted ablation studies.

\begin{table*}[t!]
\small
\centering
\begin{tabular}{cl|cccc|cccc}
\cline{2-10}
                             &                      & \multicolumn{4}{c|}{Yelp}                 & \multicolumn{4}{c}{Amazon}               \\ 
\cline{2-10}
                             & Model                & R-1 & R-2 & R-L\phantom{*} & $\mathrm{F}_{\mathrm{BERT}}$\phantom{*} & R-1\phantom{*} & R-2\phantom{*} & R-L & $\mathrm{F}_{\mathrm{BERT}}$ \\
\cline{2-10}
\multirow{4}{*}{\rot{\footnotesize Extractive}}
                             & Clustroid~{\scriptsize \citep{Brazinskas2020Unsupervised}}& 26.28   & 3.48    & 15.36\phantom{*}   & 85.8\phantom{*}            & 29.27\phantom{*}   & 4.41\phantom{*}    & 17.78   & 86.4            \\
                             & Lead~{\scriptsize\citep{Brazinskas2020Unsupervised}}      & 26.34   & 3.72    & 13.86\phantom{*}   & 85.1\phantom{*}            & 30.32\phantom{*}   & 5.85\phantom{*}    & 15.96   & 85.8            \\
                             & Random~{\scriptsize\citep{Brazinskas2020Unsupervised}}    & 23.04   & 2.44    & 13.44\phantom{*}   & 85.1\phantom{*}            & 28.93\phantom{*}   & 4.58\phantom{*}    & 16.76   & 86.0            \\
                             & LexRank~{\scriptsize\citep{Erkan2004Lexrank}}             & 24.90   & 2.76    & 14.28\phantom{*}   & 85.4\phantom{*}            & 29.46\phantom{*}   & 5.53\phantom{*}   & 17.74  & 86.4            \\
\cline{2-10}
\multirow{6}{*}{\rot{\footnotesize Abstractive}}
                             & Opinosis~{\scriptsize\citep{Ganesan2010Opinosis}}         & 20.62   & 2.18    & 12.55\phantom{*}   & 84.4\phantom{*}                                  & 24.04\phantom{*}  & 3.69\phantom{*}   & 14.58  & 85.2            \\
                             & MeanSum~{\scriptsize\citep{Chu2019Meansum}}               & 28.86   & 3.66    & 15.91\phantom{*}   & 86.5\phantom{*}                                   & 29.20\phantom{*}   & 4.70\phantom{*}    & 18.15   & -           \\
                             & DenoiseSum~{\scriptsize\citep{Amplayo2020Unsupervised}}   & 30.14   & 4.99    & 17.65\phantom{*}   & 85.9\phantom{*}                                   & -       & -       & -       &  -         \\
                             & Copycat~{\scriptsize\citep{Brazinskas2020Unsupervised}}   & 29.47   & 5.26    & 18.09\phantom{*}   & \underline{87.4}\phantom{*}                       & \underline{31.97}\phantom{*}   & \underline{5.81}\phantom{*}    & \underline{20.16}   & \underline{87.7}            \\
                             & Self \& Control~{\scriptsize\citep{Elsahar2020Self}}      & \underline{32.76} & \textbf{8.65}      & \underline{18.82}\phantom{*} & 86.8\phantom{*}    & -       & -       & -       &  -         \\
                             & MultimodalSum (ours)                                                      & \textbf{33.00}    & \underline{6.63}   & \textbf{19.84}*              & \textbf{87.7}*     &\textbf{34.19}*  &\textbf{7.05}* &\textbf{20.81} & \textbf{87.9}         \\
\cline{2-10}
\end{tabular}
\caption{Opinion summarization results on Yelp and Amazon datasets. R-1, R-2, R-L, and $\mathrm{F}_{\mathrm{BERT}}$ refer to ROUGE-\{1,2,L\}, and BERT-score, respectively. The best models are marked in bold, and the second-best models are underlined. * indicates that our model shows significant gains ($p < 0.05$) over the second-best model based on paired bootstrap resampling~\citep{Koehn2014Statistical}. All the reported scores are based on F1.}
\label{tab:summarization_results}
\end{table*}

\subsection{Main Results}

\begin{table}[t!]
\footnotesize
\centering
\begin{tabularx}{0.48\textwidth}{X}
\hline 
\multicolumn{1}{c}{Gold} \\ \hline
Wow, where to start? Some of the best sweet foods I've ever had. I wasn't sure what to try, so I tried a few things, and oh my goodness they were delicious. That's not all though, they serve drinks too so I got a latte and that was good too. There is a lot of variety here to choose from that'll make any sweet tooth salivate. Definitely a good place!\\ \hline
\multicolumn{1}{c}{Copycat} \\ \hline
If you're looking for a sweet tooth this is the place to go if you want a delicious dessert. I had the lemon meringue pie and it was delicious. The only thing I didn't like was that I could eat half of it, but it was a little pricey for what you get. \\ \hline
\multicolumn{1}{c}{Self \& Control}\\ \hline
If you're a fan of the Matos Buffet, this is a good place to visit. If you want to have a sweet tooth this is the place to be. The desserts are delicious and they have a good variety of desserts to choose from. The only thing I don't like about this place is that you have to wait in line for a long time to get in. Other than that, you can't really go wrong with any of the desserts in the buffet. The drinks are good and the desserts are yummy too. They also have desserts that are not too sweet. I'm not a huge fan of buffets, but this is one of my favorite buffets. \\ \hline
\multicolumn{1}{c}{MultimodalSum} \\ \hline
This is a cute little bakery located in the M resort. I had the chocolate croissant and it was very good. The croissants were soft and moist and the filling was delicious. I also had a chocolate chip cookie which was also good. I would definitely recommend this place if you are in the area. \\ \hline
\end{tabularx}
\caption{Sample summaries generated by various models on the Yelp dataset}
\label{tab:sample_summaries}
\end{table}

\subsubsection{Automatic Evaluation}

To evaluate the summarization quality, we used two automatic measures: ROUGE-\{1,2,L\}~\citep{Lin2004Rouge} and BERT-score~\citep{Zhang2019Bertscore}.
The former is a token-level measure for comparing $1$, $2$, and adaptive L-gram matching tokens, and the latter is a document-level measure using pretrained BERT~\citep{Devlin2019Bert}. Contrary to ROUGE-score, which is based on exact matching between n-gram words, BERT-score is based on the semantic similarity between word embeddings that reflect the context of the document through BERT. It is approved that BERT-score is more robust to adversarial examples and correlates better with human judgments compared to other measures for machine translation and image captioning. We hypothesize that BERT-score is strong in opinion summarization as well, and BERT-score would complement ROUGE-score.

The results for opinion summarization on two datasets are shown in Table~\ref{tab:summarization_results}. 
MultimodalSum showed superior results compared with extractive and abstractive baselines for both token-level and document-level measures. From the results, we conclude that the multimodal framework outperformed the unimodal framework for unsupervised opinion summarization.
In particular, our model achieved state-of-the-art results on the Amazon dataset and outperformed the comparable model by a large margin in the R-L representing the ROUGE scores on the Yelp dataset.
Although Self \& Control showed high R-2 score, we attributed their score to the inferred $N$-gram control tokens used as additional inputs to the text decoder.

Sample summaries on the Yelp dataset are shown in Table~\ref{tab:sample_summaries}.
They were generated from source reviews on Baby Cakes bakery. Copycat misused ``sweet tooth'' and generated ``lemon mernigue pie'' that was not mentioned in the source reviews. Self \& Control generated a summary about a buffet by totally misunderstanding one sentence from source reviews: ``If you love the desserts in Studio B Buffet in the M Hotel but don't want to wait in the massive buffet line or even eat in the buffet, Baby Cakes in the M Hotel is really nice fix.'' Furthermore, ``Matos Buffet'' is a non-existent word. On the contrary, MultimodalSum generated a good summary with a rich description of chocolate croissants. Although ``chocolate chip cookie'' was not found in the source reviews, our model generated it from cookie images. Note that the term can be found in other reviews that were not used as source reviews. Additional sample summaries on two datasets are shown in Appendix~\ref{appen:real_examples}.

\begin{figure*}[t!]
     \centering
     \includegraphics[width=0.95\textwidth]{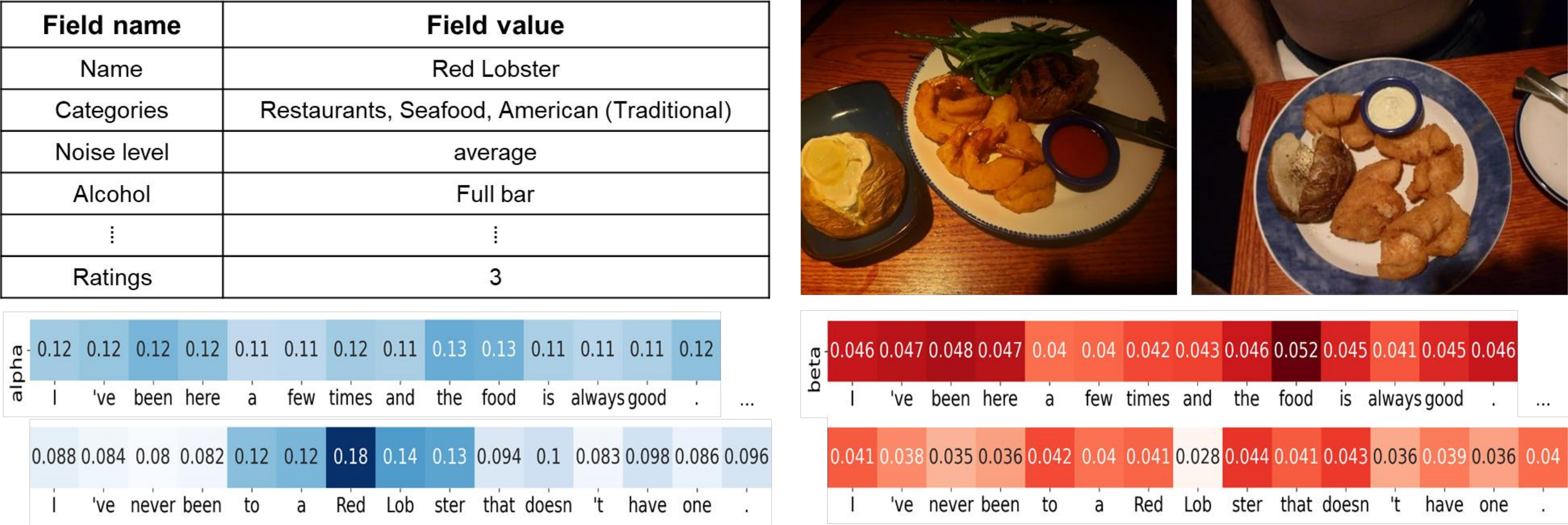}
     \caption{Multimodal gate heatmaps; From the table and two images, our model generates a summary. Heatmaps represent the overall influence of table and images for generating each word in the summary. Note that the summary is a real example generated from our model without beam search.}
     \label{fig:qualitative_modality_gate}
\end{figure*}

\subsubsection{Human Evaluation}

To evaluate the quality of summarization based on human criteria, we conducted a user study. We assessed the quality of summaries using Best-Worst Scaling (BWS; \citet{Louviere2015Best}). BWS is known to produce more reliable results than raking scales~\citep{Kiritchenko2017Best} and is widely used in self-supervised opinion summarization studies. We recruited $10$ NLP experts and asked each participant to choose one best and one worst summary from four summaries for three criteria. For each participant's response, the best model received +$1$, the worst model received -$1$, and the rest of the models received $0$ scores. The final scores were obtained by averaging the scores of all the responses from all participants.

For \textit{Overall} criterion, Self \& Control, Copycat, MultimodalSum, and gold summaries scored -$0.527$, -$0.113$, +$0.260$, and +$0.380$ on the Yelp dataset, respectively. MultimodalSum showed superior performance in human evaluation as well as automatic evaluation. We note that human judgments correlate better with BERT-score than ROUGE-score. Self \& Control achieved a very low human evaluation score despite its high ROUGE-score in automatic evaluation. We analyzed the summaries of Self \& Control, and we found several flaws such as redundant words, ungrammatical expressions, and factual hallucinations. It generated a non-existent word by combining several subwords. It was particularly noticeable when a proper noun was generated. Furthermore, Self \& Control generated an implausible sentence by copying some words from source reviews. From the results, we conclude that both automatic evaluation and human evaluation performances should be supported to be a good summarization model and BERT-score can complement ROUGE-score in automatic evaluation. Details on human evaluation and full results can be found in Appendix~\ref{appen:human_eval}.

\subsubsection{Effects of Multimodality}

To analyze the effects of multimodal data on opinion summarization, we analyzed the multimodal gate. Since the multimodal gate is a $e_D$-dimensional vector, we averaged it by a scalar value. Furthermore, as multimodal gates exist for each layer of the text decoder, we averaged them to measure the overall influence of a table or images when generating each token in the decoder.
An example of aggregated multimodal gates is shown in Figure~\ref{fig:qualitative_modality_gate}. It shows the table and images used for generating a summary text, and the multimodal gates for a part of the generated summary are expressed as heatmaps. 
As we intended, table and image information was selectively used to generate a specific word in the summary. The aggregated value of the table was relatively high for generating ``Red Lobster'', which is the name of the restaurants.
It was relatively high for images, when generating ``food'' that is depicted in two images.
Another characteristic of the result is that aggregated values of the table were higher than those of the image: mean values for the table and image in the entire test data were 0.103 and 0.045, respectively. This implies that table information is more used when creating a summary, and this observation is valid in that the table contains a large amount of metadata. Note that the values displayed on the heatmaps are small by and large, as they were aggregated from $e_D$-dimensional vector.

\subsection{Ablation Studies}

For ablation studies, we analyzed the effectiveness of our model framework and model training pipeline in Table~\ref{tab:ablation_studies}.
To analyze the model framework, we first compared the summarization quality with four versions of unimodal model framework, as in the first block of Table~\ref{tab:ablation_studies}.
BART denotes the model framework in Figure~\ref{fig:main1}, whose weights are the weights of BART-Large.
It represents the lower bound of our model framework without any training.
BART-Review denotes the model framework whose weights are from further pretrained BART using the entire training review corpus. 
UnimodalSum refers to the results of the text modality pretraining, and we classified it into two frameworks according to the use of the rating deviation.

Surprisingly, using only BART achieved comparable or better results than many extractive and abstractive baselines in Table~\ref{tab:summarization_results}.
Furthermore, further pretraining using the review corpus brought performance improvements. Qualitatively, BART with further pretraining generated more diverse words and rich expressions from the review corpus. This proved our assumption that denoising autoencoder-based pretraining helps in self-supervised multimodal opinion summarization.
Based on the BART-Review, UnimodalSum achieved superior results. 
Furthermore, the use of rating deviation improved the quality of summarization. We conclude that learning to generate reviews based on wide ranges of rating deviations including $0$ during training helps to generate a better summary of the average semantics of the input reviews. 

To analyze the effect of other modalities in our model framework, we compared the summarization quality with three versions of multimodal model frameworks, as in the second block of Table~\ref{tab:ablation_studies}. We removed the image or table modality from MultimodalSum to analyze the contribution of each modality. Results showed that both modalities improved the summarization quality compared with UnimodalSum, and they brought additional improvements when used altogether. This indicates that using non-text information helps in self-supervised opinion summarization. As expected, the utility of the table modality was higher than that of the image modality. The image modality contains detailed information not revealed in the table modality (e.g., appearance of food, inside/outside mood of business, design of product, and color/texture of product). However, the information is unorganized to the extent that the utility of the image modality depends on the capacity of the image encoder to extract unorganized information. 
Although MultimodalSum used a representative image encoder because our study is the first work on multimodal opinion summarization, we expect that the utility of the image modality will be greater if unorganized information can be extracted effectively from the image using advanced image encoders.

For analyzing the model training pipeline, we removed text modality or/and other modalities pretraining from the pipeline. By removing each of them, the performance of MultimodalSum declined, and removing all of the pretraining steps caused an additional performance drop.
Although MultimodalSum without other modalities pretraining has the capability of text summarization, it showed low summarization performance at the beginning of the training due to the heterogeneity of the three modality representations. However, MultimodalSum without text modality pretraining, whose image and table encoders were pretrained using BART-Review as a pivot, showed stable performance from the beginning, but the performance did not improve significantly.
From the results, we conclude that both text modality and other modalities pretraining help the training of multimodal framework. For the other modalities pretraining, we conducted a further analysis in the Appendix~\ref{appen:effects}.

\begin{table}[t!]
\small
\begin{center}
\begin{tabular}{l|ccc}
\hline
Models                             & R-L   \\ \hline   
BART                               & 14.85 \\       
BART-Review                        & 15.23 \\        
UnimodalSum w/o rating deviation     & 18.98 \\       
UnimodalSum w/ rating deviation       & 19.40 \\ \hline       
MultimodalSum                                 & 19.84 \\   
\quad w/o image modality               & 19.54 \\
\quad w/o table modality               & 19.47 \\ \hline
\quad w/o other modalities pretraining & 19.26 \\       
\quad w/o text modality pretraining    & 19.24 \\        
\quad w/o all modalities pretraining   & 19.14 \\ \hline 
\end{tabular}
\end{center}
\vspace{-4mm}
\caption{Ablation studies on the Yelp dataset. The first and second blocks represent various versions of the unimodal model framework and multimodal model framework, respectively. The third block shows the differences in our multimodal framework's performance according to the absence of specific steps in the model training pipeline.}
\label{tab:ablation_studies}
\end{table}

\section{Conclusions}

We proposed the first self-supervised multimodal opinion summarization framework.
Our framework can reflect text, images, and metadata together as an extension of the existing self-supervised opinion summarization framework.
To resolve the heterogeneity of multimodal data, we also proposed a multimodal training pipeline.
We verified the effectiveness of our multimodal framework and training pipeline with various experiments on real review datasets.
Self-supervised multimodal opinion summarization can be used in various ways in the future, such as providing a multimodal summary or enabling a multimodal retrieval.
By retrieving reviews related to a specific image or metadata, controlled opinion summarization will be possible.

\section*{Acknowledgments}

We thank the anonymous reviewers for their insightful comments and suggestions.

\clearpage
\bibliographystyle{acl_natbib}

\clearpage
\appendix

\section{Appendix}

\subsection{Dataset Preprocessing} \label{appen:dataset}

We selected businesses and products with a minimum of 10 reviews and popular entities above the $90^{th}$ percentile were removed. The minimum and maximum length of the words were set as $35$ and $100$ for Yelp, and $45$ and $70$ for Amazon, respectively. We set the maximum number of tokens as $128$ using the BART tokenizer for training, and we did not limit the maximum tokens for inference. For the Amazon dataset, we selected 4 categories: Electronics; Clothing, Shoes and Jewelry; Home and Kitchen; Health and Personal Care.
As Yelp dataset contains unlimited number of images for each entity, we did not use images for popular entities above the $90^{th}$ percentile. On the other hand, Amazon dataset contains a single image for each entity. Therefore, we did not use images only when meaningless images such as non-image icon or update icon were used or the image links had expired.

For Yelp dataset, we selected name, ratings, categories, hours, and attributes among the metadata. We used the hours of each day of the week as seven fields and used all metadata contained in attributes as each field. For some attributes (`Ambience', `BusinessParking', `GoodForMeal') that have subordinate attributes, we used each subordinate attribute. Among the fields, we selected $47$ fields used by at least $10\%$ of the entities. We set the maximum number of categories as $6$ based on the $90^{th}$ percentile, and averaged the representations of each category.
For ratings, we converted it to binary notation consisting of $4$ digits ($2^2, 2^1, 2^0, 2^{-1}$).
For hours, we considered (open hour, close hour) as a $2$-dimensional vector, and conducted $K$-means clustering. We selected four clusters based on silhouette score: ($16.5, 23.2$), ($8.7, 17.1$), ($6.4, 23$), and ($10.6, 22.6$). Based on the clusters, we converted hours into a categorical type.

For Amazon dataset, we selected six fields: name, price, brand, categories, ratings, and description. We set the maximum number of categories as $3$ based on the $90^{th}$ percentile, and averaged the representations of each category.
Furthermore, as each category consists of hierarchies with a maximum of 8 depths, we averaged the representations of hierarchies to get each category representation.
For price and ratings, we converted them to binary notation consisting of $11$ and $4$ digits, respectively, after rounding them to the nearest $0.5$ to contain digit for $2^{-1}$.
As some descriptions consist of many tokens, we set the maximum number of tokens as $128$. We regarded each token in description as each field, so we got total $5+128$ fields.

\subsection{Experimental Details} \label{appen:experimental}

Our image encoder is based on ResNet101. ResNet101 is composed of 1 convolution layer, 4 convolution layer blocks, and 1 fully connected layer block. Among them, 4 convolution layer blocks play an important role in analyzing image. Through each convolution layer block, the size of the image feature map is reduced to 1/4, but it gets high-level features. To maintain the ability to extract low-level features of the image, we set the model weights up to the second convolution layer block not to be trained further. We only used up to the third convolution layer block to increase the resolution of feature maps without using too high-level features for image classification. In this way, $l_I$ was set to $14\times14$ and $e_I$ was set to 1,024. 

To use the knowledge of text modality in table encoder, we obtained field name embeddings by summing the BART token embeddings for the tokens contained in the field name. Because various data types can be used for field value, we used different processing methods for each data type. Nominal values were handled in the same way as the field name. Binary and ordinal values were processed by replacing them with nominal values of corresponding meanings: `true' and `false' were used for binary values, and `cheap', `average', `expensive', and `very expensive' were used for `RestaurantsPriceRange'. Numerical values were converted to binary notation, and we obtained the representations by summing embeddings corresponding to the place, where the place value is 1. For other categorical values, we simply trained embeddings corresponding to each category.

\begin{table}[t!]
\small
\begin{center}
\begin{tabular}{l|cccc}
\hline
Pipeline step    & batch & epochs & warmup & lr    \\ \hline
Text pretrain   & 16    &  5     & 0.5    & 5e-05 \\  
Others pretrain & 32    & 20     &  1     & 1e-04 \\ 
Multimodal train &  8    &  5     & 0.25   & 1e-05 \\ \hline
\end{tabular}
\end{center}
\vspace{-4mm}
\caption{Hyperparameter values for each step in model training pipeline.}
\label{tab:hyperparameters_pipeline}
\end{table}

We set each hyperparameter value different for each step in the model training pipeline, as in Table~\ref{tab:hyperparameters_pipeline}.
We set the batch size according to the memory usage and set other values according to the amount of learning required. Hyperparameter ranges for epochs and lr (learning rate) were [3, 5, 10, 15, 20] and [1e-03, 1e-04, 5e-05, 1e-05, 5e-06], respectively, and optimized values were chosen from validation loss in one trial.
For summary generation at test time, we set different hyperparameter values for each dataset. Beam size, length penalty, and max length were set to $4$, $0.97$, and $105$ for Yelp and $2$, $0.9$, and $80$ for Amazon, respectively. Note that max length was set first to prevent incomplete termination and length penalty was determined based on the ROUGE scores on validation dataset.
The number of training parameters for text, image, and table modality pretraining are $406.3$M, $27.1$M, and $3.2$M, respectively, and that for multimodal training is $486.9$M. Run time for text modality pretraining was $16$h on $4$ GPUs, and it took $41$h and $43$h on $2$ GPUs for image and table modality training, respectively. For final multimodal training, it took $14$h on $8$ GPUs.

\subsection{Human Evaluation} \label{appen:human_eval}

\begin{table}[t!]
\scriptsize
\begin{center}
\begin{tabular}{l|ccc}
\hline
Models               & Grammaticality            & Coherence                 & Overall      \\ \hline
Self \& Control      & -0.517                    & -0.500                    & -0.527        \\ 
Copycat              & \phantom{-}0.163          & -0.077                    & -0.113        \\
MultimodalSum       & \phantom{-}\textbf{0.367} & \phantom{-}\textbf{0.290} & \phantom{-}\textbf{0.260}         \\ \hline
Gold                 & -0.013                    & \phantom{-}0.287          & \phantom{-}0.380\\ \hline
\end{tabular}
\end{center}
\vspace{-4mm}
\caption{Human evaluation results in terms of the BWS on the Yelp dataset.}
\label{tab:human_evaluation}
\end{table}

For human evaluation, we randomly selected $30$ entities from Yelp test data, and used three criteria: \textit{Grmmaticality} (the summary should be fluent and grammatical), \textit{Coherence} (the summary should be well structured and well organized), and \textit{Overall} (based on your own criteria, select the best and the worst summary of the reviews).
Results for three criteria are shown in Table~\ref{tab:human_evaluation}. 
Self \& Control achieved very poor performance for all criteria due to its flaws that were not revealed in the automatic evaluation. 
Surprisingly, MultimodalSum outperformed gold summaries for two criteria; however, its overall performance lagged behind Gold. As our model was initialized from BART-Large that had been pretrained using large corpus and further pretrained using training review corpus, it may have generated fluent and coherent summaries. It seems that our model lagged behind Gold in \textit{Overall} due to various criteria other than those two. The fact that Gold scored lower than Copycat in \textit{Grammaticality} may seem inconsistent with the result from \citet{Brazinskas2020Unsupervised}. However, we assumed that this result was due to a combination of the four models in relative evaluation. The ranking for Copycat and Gold may have changed in absolute evaluation.

\subsection{Analysis on Other Modalities Pretraining} \label{appen:effects}

To analyze the various models for the other modalities pretraining, we evaluated the performance of the reference review generation task that generates corresponding reviews from images or a table.
For evaluation, we used the data that were not used for training data: we left $10\%$ of the data for Yelp and $5\%$ for Amazon.
We chose two comparison models: Untrained and Triplet.
Untrained denotes the model that image encoder or table encoder keeps untrained.
This option indicates the lower bound containing only the effect of the text decoder.
Triplet denotes the triplet-based metric-learning model, based on \citet{Lee2018Stacked} and \citet{Vo2016Localizing}.
For triplet (images or a table, reviews of positive entity, reviews of negative entities), we trained the image or table encoder based on the pretrained text encoder, by placing the image or table encoded representations close to the positive reviews representations and far from the negative reviews representations. Note that pretrained text encoder was not trained further.

\begin{table}[t!]
\scriptsize
\begin{center}
\begin{tabular}{l|ccc|ccc}
\hline
                     & \multicolumn{3}{c|}{Image}  & \multicolumn{3}{c}{Table} \\ \hline
Models               & R-1    & R-2   & R-L       & R-1    & R-2   & R-L             \\ \hline
Untrained            & 21.03 & 2.45 & 14.17    & 24.04 & 2.92 & 15.10          \\ 
Triplet              & 20.06 & 2.49 & 13.15    & 25.67 & 3.52 & 15.16          \\
Pivot (ours)                 & \textbf{25.87} & \textbf{3.62} & \textbf{15.70}    & \textbf{27.32} & \textbf{4.12} & \textbf{16.57}          \\ \hline
\end{tabular}
\end{center}
\vspace{-4mm}
\caption{Reference reviews generation results on the Yelp dataset.}
\label{tab:other_modalities_results}
\end{table}

Results on the other modalities pretraining are shown in Table~\ref{tab:other_modalities_results}.
For each model, the pretrained decoder generated a review from image or table encoded representations.
We measured the average ROUGE scores between the generated review and $N$ reference reviews.
The first finding was that results of table outperformed those of image. It indicates that table has more helpful information for generating reference review.
The second finding was that our method based on the text decoder outperformed the Triplet based on the text encoder. Especially, Triplet achieved very poor performance for image because it is hard to match $M$ images to $N$ reference reviews for metric learning. On the contrary, our method achieved much better performance by pivoting the text decoder.
Triplet showed good performance on table because it is relatively easy to match $1$ table to $N$ reference reviews; however, our method outperformed it. We conclude that our method lets the image and table encoder get proper representations to generate reference reviews regardless of the number of inputs.

\clearpage
\onecolumn
\subsection{Example Summaries} \label{appen:real_examples}

Table~\ref{tab:yelp_summaries}, \ref{tab:amazon_summaries} show sample summaries generated from our model and baseline models on Yelp and Amazon datasets. Full summaries from our model are available at \url{https://bit.ly/3bR4yod}.

\begin{table*}[h!]
\small

\includegraphics[width=\linewidth, height=35mm]{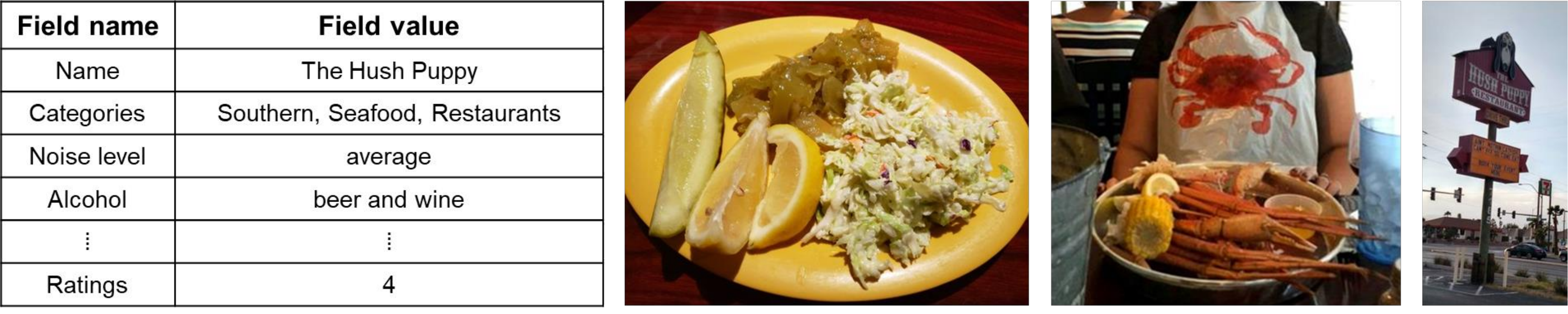}
\vspace{0.1cm}

\renewcommand\tabularxcolumn[1]{m{#1}}
\begin{tabularx}{\textwidth}{|c|X|}
\hline
Review 1 & The fresh water catfish is probably the best I've every had. The service was outstanding. I would recommend this little secret to everyone. \\ \hline
Review 2 & I loved everything about this place!! Great food, great decor, and great service. The best collard greens I have ever had. We had fried oysters for a starter and although I have never had them before so I have nothing to compare them with they were very tasty. The warm hush puppies with the honey butter was delicious!! I had the crab legs which were perfect and plentiful. My sister had the all you can eat fried catfish that was also cooked perfectly. A great experience all around!! \\ \hline
Review 3 & Amazing food and great service! The hospitality was out of this world. Will definitely be back soon. The wait was less than 5 minutes at 7pm on a Friday night, amazing!! The staff was very kind and the waitresses were very attentive and helpful. We tried the frog legs, catfish, alligator bites, crab legs, gumbo and of course the hush puppies! Everything was outstanding. What a hidden gem! \\ \hline
Review 4 & I love this place the food amazing the staff helpful ....must try green tomatos ...fresh water fish ;\char`\^ ) \\ \hline
Review 5 & We love this place the catfish is good the hush puppies with that honey butter are awesome the french fries the gumbo what else is good there the alligator tail mostly everything on the menu. I guess the only bad thing I can say is sometimes it's like a 20 minute wait in the drive-through but it's well worth it when your food is hot Because tonight I got to go home and warm it up it's not hot enough, Even though they're still open for another hour that was a bummer \\ \hline
Review 6 & Really tasty catfish, shrimp and fixin's. Our friend took us to the sister location on Nellis a couple of months ago, but this location was more convenient to our hotel. No worries, this place was just as good! Excellent service, and the salad bar is a nice touch as well. As a Bostonian, I'm pretty particular about seafood. The Hush Puppy fits the bill. Very satisfied! \\ \hline
Review 7 & First Time here and the food, staff was awesome. Manager came over and gave us samples of the fried catfish, super nice. \\ \hline
Review 8 & I never eat catfish. It's nasty to me until I tasted the saltwater catfish!!! Greens are on point. The hushpuppy are bomb with honey butter!!!!! Gator bites where are ok. \\ \hline \hline
Copycat & This place is awesome! The food was great, the service was great. We had the catfish po'boy and it was delicious. The only reason I didn't give 5 stars is because of the fact that they don't deliver.\\ \hline
Self \& Control & I love this place. The service is awesome. The hush puppies are to die for. I love the honey butter. I can't wait to go back and try it again. The only thing I don't like about the place is the wait. It can be a little long, but it's worth it. It's a little on the pricey side, but you're getting what you pay for. Love the hot butter, the hush puppies, the French fries, the gumbo, the catfish and the gumbo. Everything is so yummy and the service is top notch. Try it out, you won't be disappointed.\\ \hline
MultimodalSum & This place is a hidden gem. The food is great and the service is even better. I had the all you can eat catfish and it was delicious. The hush puppies are the best I've ever had. I will definitely be back.\\ \hline
Gold & Yummy and delicious catfish. You gotta try it. Friendly staff and service is good too. You can tell they know their seafood and how to prepare and cook it to perfection. The staff also answered any questions I had. The Hush Puppies are tasty too.\\ \hline
\end{tabularx}
\caption{Yelp summaries generated by different models.}
\label{tab:yelp_summaries}
\end{table*}

\clearpage
\begin{table*}[t]
\small

\includegraphics[width=\linewidth, height=40mm]{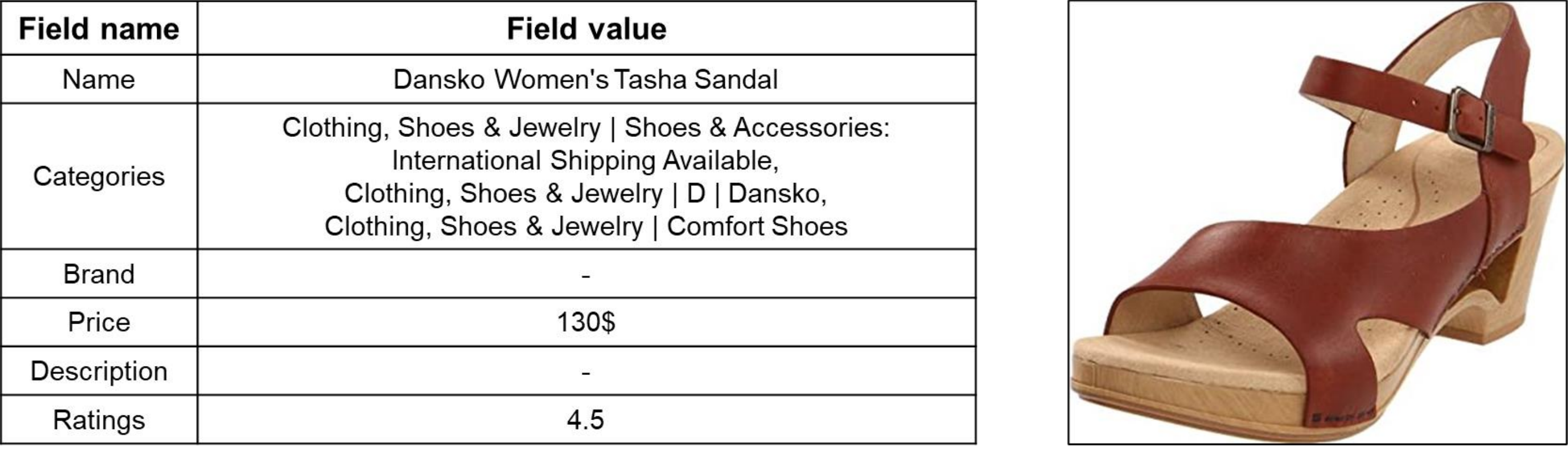}
\vspace{0.1cm}

\renewcommand\tabularxcolumn[1]{m{#1}}
\begin{tabularx}{\textwidth}{|c|X|}

\hline
Review 1 & I usually wear size 37, but found a 38 feels better in this sandal. I absolutely love this sandal. So supportive and comfortable, although at first I did get a blister on my big toe. Do not let this be the deciding factor. It stretched out and is now fabulous. I love it so much that I bought it in three colors. \\ \hline
Review 2 & This is a really cute shoe that feels very comfortable on my high arches. The strap on the instep fits my feet very well, but I have very slim feet. I can see how it would be uncomfortably tight on anyone with more padding on their feet. \\ \hline
Review 3 & I love these sandals. The fit is perfect for my foot, with perfect arch support. I don't think the leather is cheap, and the sandals are very comfortable to walk in. They are very pretty, and pair very well with pants and dresses. \\ \hline
Review 4 & My wife is a nurse and wears dansko shoes. We were excited to try the new crimson sandal and normally order 39 sandal and 40 closed toe. Some other reviews were right about a narrow width and tight toe box. We gave them a try and passed a great pair of shoes to our daughter with her long narrow feet, and she loves them... \\ \hline
Review 5 & Finally, a Dansko sandal that's fashion forward! It was love at first sight! This is my 4th Dansko purchase. Their sizing, quality and comfort is very consistent. I love the stying of this sandal and I'm pleased they are offering bolder colors. Another feature I love is the Dri-Lex topsole - it's soft and keeps feet dry. \\ \hline
Review 6 & I really love these sandals. my only issue is after wearing them for a while my feet started to swell as I have a high instep and they were a little tight across the top. I'm sure they will stretch a bit after a few wears \\ \hline
Review 7 & I have several pairs of Dansko clogs that are all size 39 and fit perfectly. So I felt confident when I ordered the Tasha Sandal in size 39. I don't know if a 40 would be too large but the 39 seems a little small. Otherwise, I love them. They are very cushiony and comfortable! \\ \hline
Review 8 & I own many Dansko shoes and these are among my favorites. They have ALL the support that Dansko offers in its shoes plus they are very attractive. I love the the heel height and instant comfort. They look great with slacks and dresses, dressed up or not... \\ \hline \hline
Copycat & This is my second pair of Dansko clogs and I love them. They are very comfortable and I can wear them all day without any discomfort. I would recommend them to anyone looking for a comfortable sandal.\\ \hline
MultimodalSum & I love these sandals. They are very comfortable and look great. The only thing I don't like is that they are a little tight across the top of my foot. I have a high instep and the strap is a little too tight. I am hoping they will stretch out a bit.\\ \hline
Gold 1 & I love these sandals, Dansko has made a really great product! I had to return my first pair (39) for being a bit tight and small, but I went a size higher (40) and it is perfect, they are so comfortable! If they do stretch out like other reviews say, they will still fit and look great.\\ \hline
Gold 2 & I love these Dansko Tasha sandals! They are comfortable and the style is really cute. The only warning I have is that they seem to run narrow: you may want to buy a larger size if you have wide feet. Also, they seem to stretch as you wear them, so don't get discouraged by a few blisters on first wearing. \\ \hline
Gold 3 & These Dansko shoes are amazingly comfortable and hug the shape of my feet well, but I did have to wear them for a bit to stretch them out. They felt a little tight at first, but now they are perfect. I feel they're true to size so I'd recommend ordering these in your normal shoe size. \\ \hline
\end{tabularx}
\caption{Amazon summaries generated by different models.}
\label{tab:amazon_summaries}
\end{table*}

\end{document}